\documentclass[runningheads]{llncs}
\usepackage[T1]{fontenc}
\usepackage{graphicx}
\usepackage{cite}
\usepackage{amssymb}
\usepackage[dvipsnames]{xcolor}
\usepackage{pgfplots}
\usepgfplotslibrary{external}
\definecolor{clgray}{RGB}{87, 87, 87}
\definecolor{clblue}{RGB}{42, 75 , 215}
\definecolor{clgreen}{RGB}{29, 105, 20}
\definecolor{clbrown}{RGB}{129, 74 , 25}
\definecolor{clcyan}{RGB}{41, 208, 208}
\definecolor{clpurple}{RGB}{246, 0 , 220}
\definecolor{clyellow}{RGB}{255, 238, 51}
\definecolor{clred}{RGB}{245, 39, 39}
\definecolor{clgreenbg}{RGB}{39, 245, 39}
\definecolor{clbluebg}{RGB}{39, 39, 245}
\definecolor{clwhite}{RGB}{245, 245, 245}
\definecolor{clblack}{RGB}{10, 10, 10}

\definecolor{clpink}{RGB}{243, 139, 197}
\definecolor{clorange}{RGB}{246, 138, 0}
\definecolor{cllime}{RGB}{129, 197, 122}
\definecolor{cldgreen}{RGB}{0, 90, 41}
\definecolor{cldred}{RGB}{90, 0, 0}
\definecolor{cldblue}{RGB}{0, 28, 90}

\definecolor{cldgreenbg}{RGB}{0, 90, 41}
\definecolor{clpinkbg}{RGB}{245, 39, 245}
\definecolor{clyellowbg}{RGB}{245, 245, 39}
\definecolor{clcyanbg}{RGB}{39, 245, 245}
\newcommand{\mycirc}[1][black]{\Large\textcolor{#1}{\ensuremath\bullet}}
\begin{document}
\title{Is an object-centric representation beneficial for robotic manipulation ?}
\titlerunning{Is an OCR beneficial for learning robotic manipulation ?}
%
\author{Alexandre Chapin \and
Emmanuel Dellandrea \and
Liming Chen
}

\authorrunning{A. Chapin et al.}
%
\institute{Affiliation: Ecole Centrale de Lyon, CNRS, INSA Lyon, Université Claude Bernard Lyon 1,Université Lumière Lyon 2, LIRIS, UMR5205, 69130 Ecully, France \email{\{firstname.name\}@ec-lyon.fr}
}
\maketitle              
\begin{abstract}
Object-centric representation (OCR) has recently become a subject of interest in the computer vision community for learning a structured representation of images and videos. It has been several times presented as a potential way to improve data-efficiency and generalization capabilities to learn an agent on downstream tasks. However, most existing work only evaluates such models on scene decomposition, without any notion of reasoning over the learned representation. 
Robotic manipulation tasks generally involve multi-object environments with potential inter-object interaction. We thus argue that they are a very interesting playground to really evaluate the potential of existing object-centric work. To do so, we create several robotic manipulation tasks in simulated environments involving multiple objects (several distractors, the robot, etc.) and a high-level of randomization (object positions, colors, shapes, background, initial positions, etc.). We then evaluate one classical object-centric method across several generalization scenarios and compare its results against several state-of-the-art hollistic representations. Our results exhibit that existing methods are prone to failure in difficult scenarios involving complex scene structures, whereas object-centric methods help overcome these challenges.

\end{abstract}

\begin{figure}[h]
\begin{center}
\includegraphics[width=\linewidth]{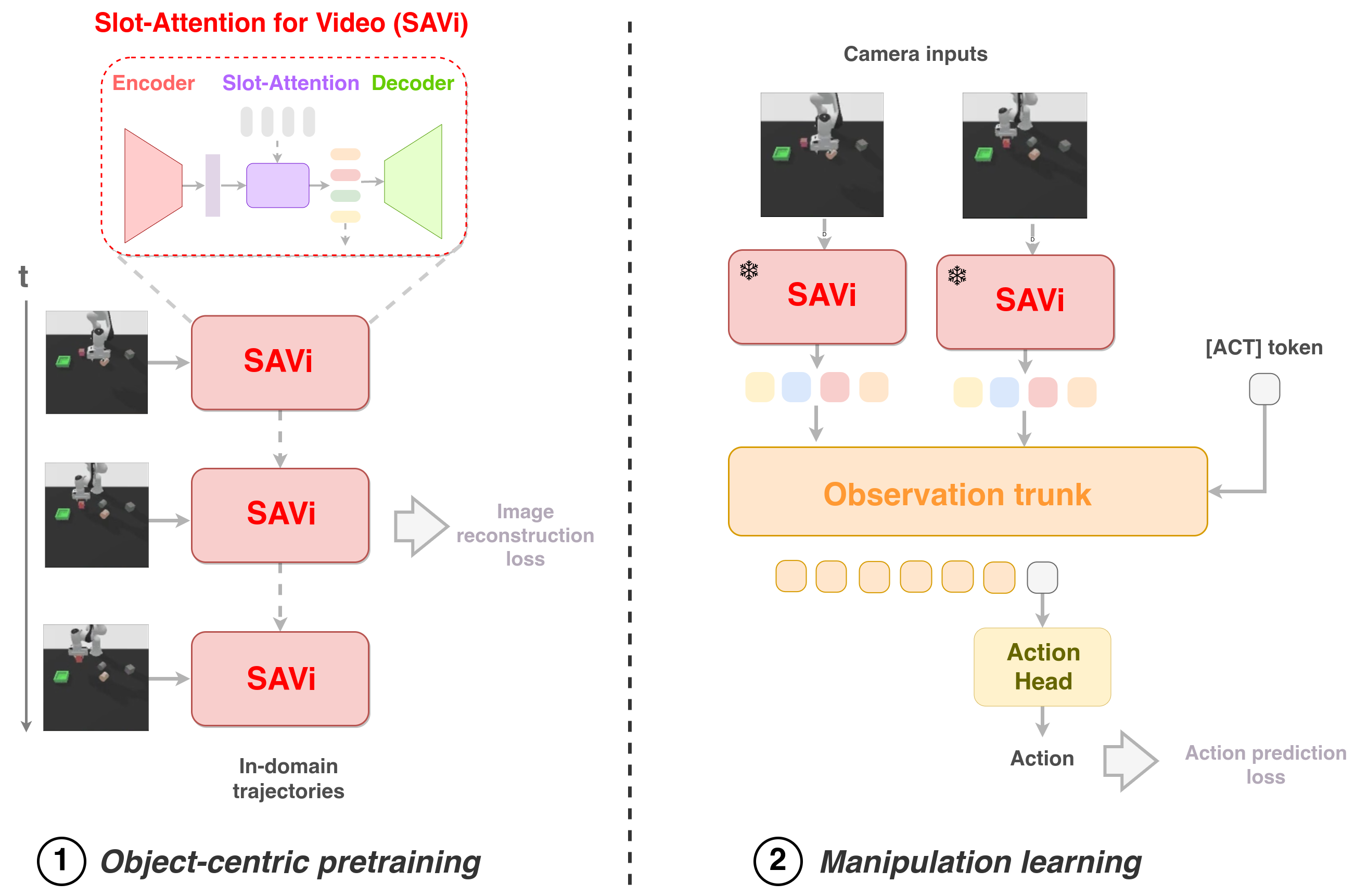}
\end{center}
\caption{\textbf{Overview of the proposed framework.} We introduce a new framework to evaluate different representations models for robot manipulation policy learning. Our main objective is to evaluate object-centric representation methods. To do so, the models can be trained following classical scheme through image reconstruction loss on in-domain data. Then, the model can be used as a frozen backbone for manipulation learning : it is used to encode camera inputs from several timesteps before being fed to a transformer models that concatenate all the information into an additional \textit{[ACT]} token. This final token is used as input to an action head in order to predict the next action. The second part of the model is learnt through imitation learning with an action prediction loss.}
\label{fig:method}
\end{figure}

\section{Introduction}
Robotic manipulation tasks generally involve a multitude of objects with numerous possible scene arrangements and interaction methods, such as pushing, grabbing, moving, lifting, stacking and more. Several challenges remain to be addressed to enable the deployment of generalizable robots in real-world settings. These challenges include predicting and understanding the dynamics between both seen and unseen objects to successfully complete the required tasks. It is well-known that one of the most crucial aspects of achieving a more efficient and generalizable robotic agent is the ability to effectively handle the representations derived from its sensors \cite{kroemer2020reviewrobotlearningmanipulation, bengio2014representationlearningreviewnew}.

In recent years, various studies have explored ways to improve the representations of the environments surrounding robots. For example, through pre-training and time-contrastive learning over human ego-centric datasets \cite{nair2022r3muniversalvisualrepresentation, majumdar2024searchartificialvisualcortex, ma2023vipuniversalvisualreward}, distilling knowledge from pre-trained computer vision (CV) models \cite{shang2024theiadistillingdiversevision}, and leveraging successful learning scheme of the CV field such as masked autoencoding \cite{radosavovic2022realworldrobotlearningmasked}. In \cite{hu2023pretrainedvisionmodelsmotor}, an extensive review and evaluation of existing state-of-the-art representation models from CV applied on robotic tasks has been made, and showed that most of existing models can be used "off-the-shelf" for policy learning. However, these approaches often represent the environment as a single global vector, which must account for the wide variety of situations a robot might encounter.

Independent of these studies, a promising approach has emerged in computer vision: object-centric representations. This method segments images into distinct entities, known as \textit{objects}, introducing structure and symbolic reasoning into the representation. While OCR's potential for robotic tasks has been acknowledged, most studies has primarily focused on image decomposition and reconstruction rather than assessing its impact on robotic applications. Some works have explored how OCR performs in scenarios such as Reinforcement Learning (RL) \cite{yoon2023investigationpretrainingobjectcentricrepresentations, heravi2023visuomotorcontrolmultiobjectscenes, haramati2024entitycentricreinforcementlearningobject}, but these are typically limited to simple 2D environments or toy tasks \cite{watters2019cobradataefficientmodelbasedrl, kipf2022conditionalobjectcentriclearningvideo, zhang2022objectcentricvideorepresentationbeneficial}.

In this work, our objective is to evaluate whether OCR improves the accuracy of the policy learning, but also the generalization capability of the final model. In the following sections, we will outline our contributions in details: 
\begin{itemize}
    \item Building on top of the existing works using transformers for behavior cloning, we setup an easily replicable transformer-based policy model using an OCR backbone as vision encoder.
    \item We build several tasks in simulated environments implying multiple objects with randomness over scenes (object, background, ...) and extract expert trajectories with motion planning to create our \textbf{RoboShape} dataset.
    \item We evaluate and compare our method against several state-of-the-art global image representation models, in order to show the effectiveness of OCR for complex robotic manipulation scenarios.
\end{itemize}

\section{Related works}
\subsection{Object-centric representation}
The learning of an OCR is a prominent and very promising field of research in CV.  Early works primarily focused on learning structured image representations through generative modeling \cite{kingma2022autoencodingvariationalbayes, higgins2017betavae}. Most methods were based on autoencoder architectures, aiming to obtain a structured and disentangled latent space. Later, some studies explored how to explicitely structure this space by decomposing the latent representation into multiple vectors, rather than using a single vector to represent the entire image \cite{burgess2019monetunsupervisedscenedecomposition, watters2019cobradataefficientmodelbasedrl, kabra2021simoneviewinvarianttemporallyabstractedobject}. One of the most significant methods in this direction is Slot-Attention \cite{locatello2020objectcentriclearningslotattention}, known for its simplicity ad efficiency, making it easy to train and deploy. Subsequent papers explored the use of more powerful decoders, such as diffusion models \cite{wu2023slotdiffusionobjectcentricgenerativemodeling, jiang2023objectcentricslotdiffusion} and transformer decoders \cite{singh2022illiteratedallelearnscompose}, and pre-trained backbones such as DINO \cite{caron2021emergingpropertiesselfsupervisedvision} to help  model learn object representation in more complex, real-world scenarios \cite{seitzer2023bridginggaprealworldobjectcentric}. Following works extended Slot-Attention to videos by incorporating recurrent neural network or transformer architectures \cite{kipf2022conditionalobjectcentriclearningvideo, elsayed2022saviendtoendobjectcentriclearning, singh2022simpleunsupervisedobjectcentriclearning, zadaianchuk2023objectcentriclearningrealworldvideos}. 

Recently, some studies have investigated the use of OCR for reinforcement learning (RL)  \cite{watters2019cobradataefficientmodelbasedrl, heravi2023visuomotorcontrolmultiobjectscenes, yoon2023investigationpretrainingobjectcentricrepresentations, haramati2024entitycentricreinforcementlearningobject}, demonstrating their potential to handle multi-entity scenes. In parallel, a transformer-based approach called SlotFormer was developed for next-frame prediction \cite{wu2023slotformerunsupervisedvisualdynamics}, predicting dynamics in the object-centric latent space. 
 This work is particularly promising as it can predict scene dynamics after unsupervised learning. Our work builds upon these advancements by proposing a framework built on top of OCR models to evaluate them in scenarios that require understanding the notion of objects. 

\subsection{Vision based imitation learning}
Imitation learning (IL) is a paradigm that allows robots to learn from expert demonstrations. One of the most prominent and simple implementations of IL is Behavior cloning (BC), which can be viewed as a form of \textit{supervised learning}. It consists of mapping observation data to actions. BC has been successful in teaching a wide range of tasks (e.g., grasping, pick and place, etc.) from low-dimensional states, demonstrating the generalizable properties of agents trained on sufficient data \cite{embodimentcollaboration2024openxembodimentroboticlearning, brohan2023rt1roboticstransformerrealworld}. Most of the work in BC has focused on creating new architectures for learning policy, ranging from simple MLPs to transformers \cite{brohan2023rt1roboticstransformerrealworld, dalal2023optimus}, diffusion models \cite{chi2024diffusionpolicyvisuomotorpolicy} and even combinations of transformers and diffusion models \cite{octomodelteam2024octoopensourcegeneralistrobot}. 
Some recent studies have explored the use of object-centric representations to learn behaviors, showing promising results on toy scenarios \cite{heravi2023visuomotorcontrolmultiobjectscenes}. 
In \cite{burns2023makespretrainedvisualrepresentations}, it was demonstrated that pre-trained models like DINO \cite{caron2021emergingpropertiesselfsupervisedvision} can provide strong generalization capabilities, even surpassing models specifically designed for robotic manipulation \cite{nair2022r3muniversalvisualrepresentation, ma2023vipuniversalvisualreward}. In \cite{qian2024recasting}, an object-centric representation was extracted from DINOv2 \cite{oquab2024dinov2learningrobustvisual}, showing that such representations not only improved the success rate of robotic manipulation over the original models but also matched the performance of state-of-the-art pre-trained robotic representation models. More recently, \cite{shang2024theiadistillingdiversevision} combined several recent state-of-the-art CV foundation models with knowledge distillation. Their final model demonstrated impressive performance across several behavior cloning benchmarks. 

In this work, we focused on developing a simple policy architecture on top of existing OCR models, which we compare with current methods for representation learning.

\section{Method}

\subsection{Preliminaries}
\paragraph{Imitation-learning}
Our aim is to learn a robotic policy capable of solving a task using vision. The robot agent receives as input a sequence of observations $(o_0, o_1, ..., o_t)$. These observations are processed by an encoder model to produce a sequence of states $(s_0, s_1, ..., s_t) = f(o_0, o_1, ..., o_t)$. Based on a fixed-horizon history $H$, the policy generates an action distribution $\pi(. \vert \{s_t\}_{t=0}^H)$ from which the action $a_{H+1}$ is sampled and applied to the robot. The entire model can be trained end-to-end or in two stages : first, pre-training the vision component on out-of-domain data, followed by freezing it for policy learning. 

\paragraph{Slot-Attention for Video (SAVi)}
One of the most prominent object-centric models for video decomposition is Slot-Attention for Video \cite{elsayed2022saviendtoendobjectcentriclearning, kipf2022conditionalobjectcentriclearningvideo}. The model consises of an encoder $f(o_t) = h_t$, typically a convolutional neural network (CNN), which processes the input image before passing it to the Slot-Attention module \cite{locatello2020objectcentriclearningslotattention}.  The Slot-Attention module clusters features into a slot representation : $SA(h_t) = \mathcal{S}_t = \{s^1, ...,s^K\} \in \mathbb{R}^{K \times D}$ where $K$ is the number of slots and $D$ is the slot size. Next, a predictor model, generally a transformer, is applied to the slots to capture temporal dynamics by propagating information between slots across timesteps: $p(\mathcal{S}_t, \mathcal{S}_{t+1}, ..., \mathcal{S}_{t+h})$, where $h$ is the temporal horizon. Finally, a decoder attempts to reconstruct the input image from the slot representation : $d(\mathcal{S}_t) = \tilde o_t$.

\subsection{Object-centric transformer policy.}

\subsubsection{Architecture}
An overview of our model is shown in Figure \ref{fig:method}, it is composed of three components : an image encoder handling the camera inputs, a transformer-based observation trunk, concatenating the informations from the observations, and finally an action head making the final decision choice. 

\paragraph{Encoder:}
Our image encoder is the video object-centric model SAVi \cite{kipf2022conditionalobjectcentriclearningvideo}. It consists of a small CNN, which is a stack of five convolutional layers with ReLU activation, concatenated with positional encoding. The resulting feature grid is flattened before being passed to the Slot-Attention process. In the Slot-Attention process, each slot is initialized as a random vector and iteratively cross-attended with the output features from the CNN. Unlike classical cross-attention, renormalization is performed over the queries instead of the keys. Each frame of the input video is encoded independently by the CNN, and the first frame is further processed by the Slot-Attention module. The output slots from this process are passed through a transformer model (predictor) and are then used as initialization slots for the subsequent frame. This step allows the model to maintain a memory of the slots over the course of the video. 

\paragraph{Observation trunk:}
To effectively handle the multiple vectors extracted from object-centric representations, we adopted the architecture proposed by \cite{haldar2024bakuefficienttransformermultitask}, which uses a transformer-based observation trunk. This approach allows for a more sophisticated integration of all information into a single vector, avoiding the limitations of traditional methods. Indeed, existing works often rely on a MLP head as a bottleneck before the action head. However, such a structure is not well-suited for object-centric representations.

The transformer takes as input the history of representation of the last $H$ frames $(\mathcal{S}_{t-H}, ..., \mathcal{S}_{t})$ with  $\mathcal{S}_t = \{s^1, ...,s^K\} \in \mathbb{R}^{K \times D}$ and an additional learnable action token $[ACT]$. The purpose of the action token is to aggregate the information from the history of slots in order to then perform the prediction. 

\paragraph{Action head:}
After a forward pass, the $[ACT]$ is isolated to predict the next action $a_t$ through the action head.
The actions are composed of 7 dimensions corresponding of the 6D relative pose of the end-effector $(x, y, z, r_x, r_y, r_z)$ and one for the gripper state $\in [-1, 1]$ (close or open). In our experiments, we use a Gaussian-Mixture Model (GMM) head, as it is done in several recent works \cite{zhu2023violaimitationlearningvisionbased, mandlekar2021matterslearningofflinehuman}, in order to handle the multi-modality of the encountered environments.  

\subsubsection{Training}
The training pipeline of our model is divided in two parts : a vision pre-training phase and a behavior learning phase. 

During the pre-training phase, the model is trained on in-domain data (the same data used for the behavior learning phase). Following the SAVi training scheme, the model processes a short input video, decomposes each frame into slots, and then reconstructs the input video from these slots. The model is learned end-to-end with an image reconstruction loss (MSE loss) and permits to obtain a good decomposition of objects over the dataset. 

In the behavior learning phase, the representation model is frozen and used to extract slot representation from the input frames. These slot representations  condition the prediction of the next action. To optimize the policy model, a negative log-likelihood loss (NLL) is applied to the predicted action distribution, comparing it to the target action. 
 
\section{Experiments}
\subsection{Experimental setup}
In order to validate our model against existing methods, we designed several multi-object robotic manipulation tasks using Maniskill \cite{tao2024maniskill3gpuparallelizedrobotics}, an open-source robotic framework built upon the SAPIEN environment \cite{Xiang_2020_SAPIEN}. In our experiments, we aim to address the following questions: 1) Can an OCR solve visual-based robotic manipulation ? 2) Can an OCR generalize better than classical global representation methods to unseen scenes, distractors, or colors  ?

\subsubsection{Tasks definition}

\begin{figure}[h]
\begin{center}
\includegraphics[width=\linewidth]{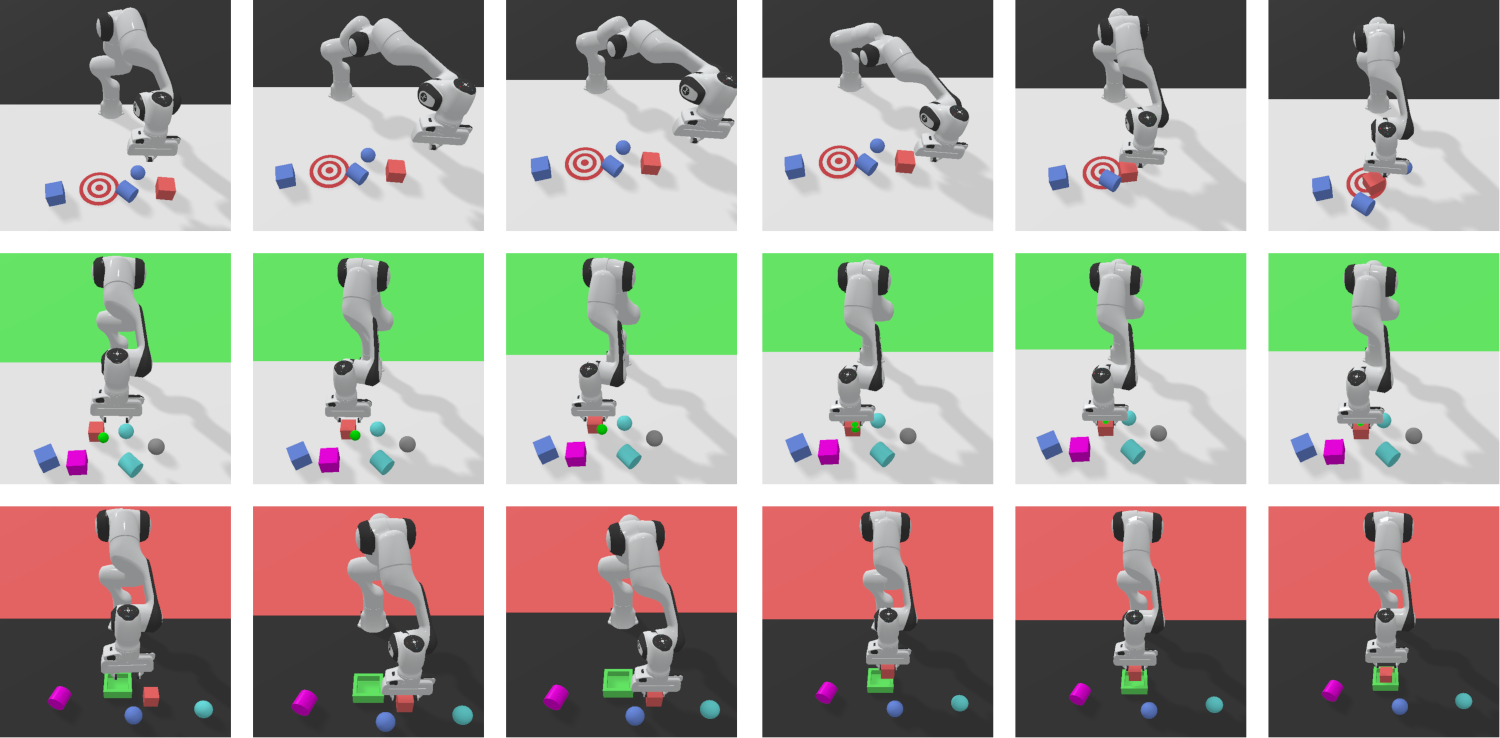}
\end{center}
\caption{
\textbf{Example rollout of the tasks.} Top row: \textit{Push cube to target}; Middle row: \textit{Pick cube}; Bottom row: \textit{Place cube in bin}}
\label{fig:tasks}
\end{figure}
The models are trained on a new dataset we created, called \textbf{RoboShape}, composed of 3 different tasks with an increasing level of difficulty as shown in Figure \ref{fig:tasks}. 
\begin{itemize}
    \item Task 1 : \textit{Push cube to target}; the goal is to push the red cube to the target.
    \item Task 2 : \textit{Pick cube}; the goal is to pick up the red cube, and place it to a pre-determined position.
    \item Task 3 : \textit{Place cube in bin}; the goal is to pick up the red cube and place it into the green bin. 
\end{itemize}

\begin{figure}[h]
\begin{center}
\includegraphics[width=\linewidth]{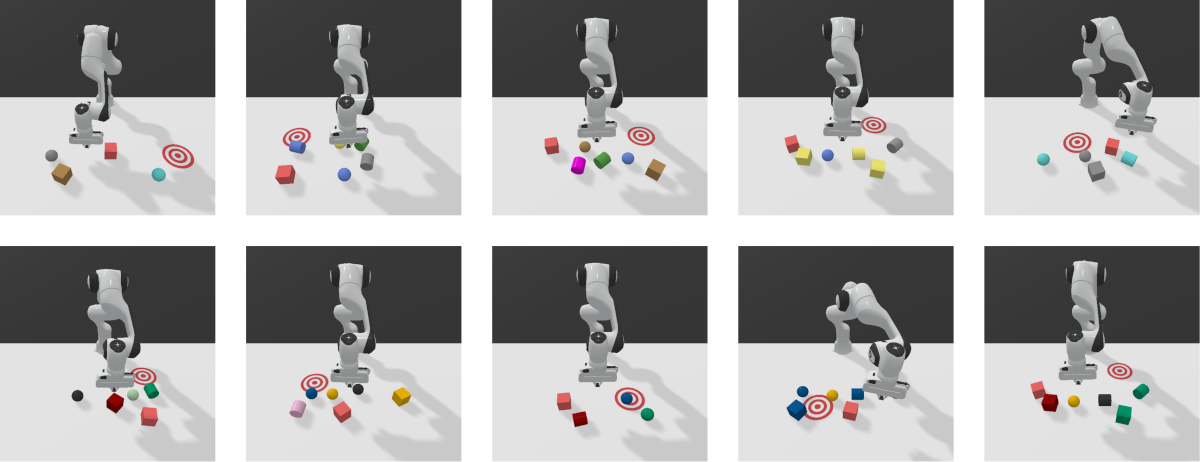}
\end{center}
\caption{
\textbf{L1 generalization - Unseen distractor colors} Top row: distractor colors seen during training; Bottom row: unseen distractor colors}
\label{fig:colors}
\end{figure}

\begin{figure}[h]
\begin{center}
\includegraphics[width=\linewidth]{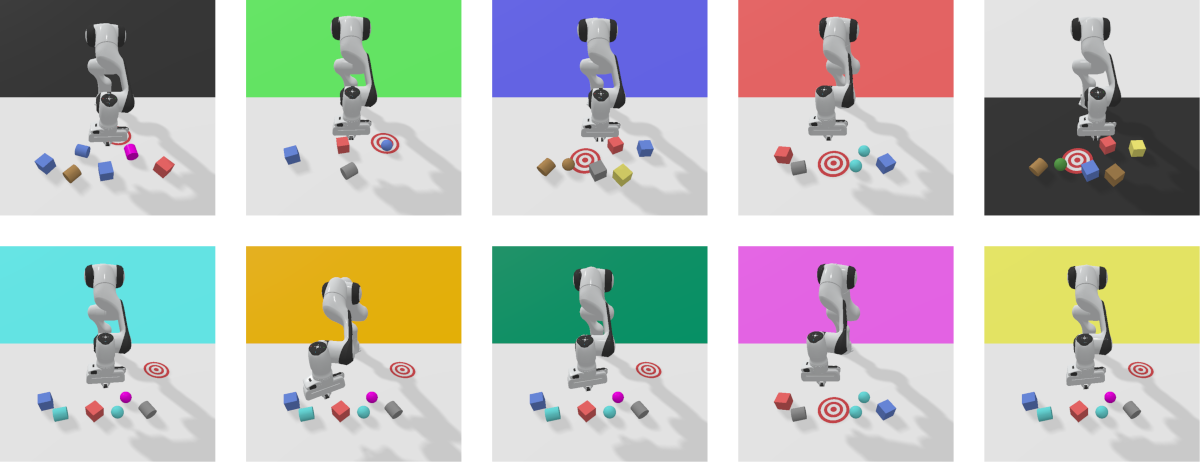}
\end{center}
\caption{
\textbf{L2 generalization - Unseen background colors} Top row: background seen during training; Bottom row: unseen background colors}
\label{fig:background}
\end{figure}

\begin{figure}[h]
\begin{center}
\includegraphics[width=\linewidth]{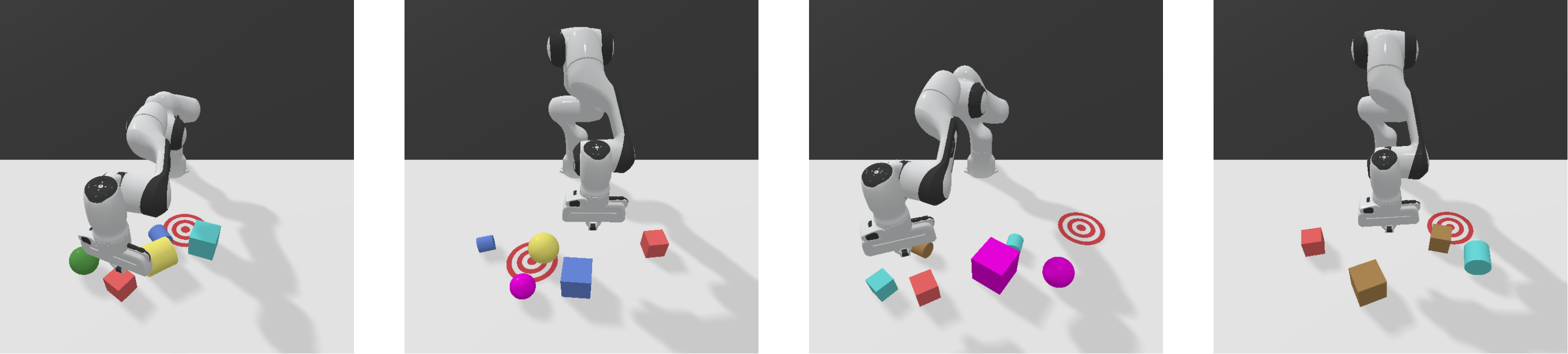}
\end{center}
\caption{
\textbf{L3 generalization - Unseen distractor sizes} the distractors have a random size and can have smaller or bigger sizes than seen during training}
\label{fig:sizes}
\end{figure}

Each task is a table-top manipulation task performed with a Franka robot using a single view positioned in front of the robot. We do not use proprioceptive information such as end-effector pose or joint positions. Using oracle information, we built a dataset of expert trajectories by collecting 2000 expert trajectories for each task. 

At each reset of the environment, several randomization processes are applied : target and distractor positions, distractor shapes and colors, background and table color and the robot's initial position. All randomization process can be observed on Figures \ref{fig:colors}, \ref{fig:background} and \ref{fig:sizes}. The colors of the distractors are defined as  [\textit{\mycirc[clgray], \mycirc[clblue], \mycirc[clgreen], \mycirc[clbrown], \mycirc[clcyan], \mycirc[clpurple], \mycirc[clyellow]}], as shown on first row of Figure \ref{fig:colors}. Those of the background are defined as [\textit{\mycirc[clred], \mycirc[clgreenbg], \mycirc[clbluebg], \mycirc[clwhite], \mycirc[clblack]}], as shown on first row of Figure \ref{fig:background}. 

Given the high level of randomization, especially on the target positions, the policy needs to understand high-level concepts in order to solve the tasks such as "what are the objects I need to interact with ?", "where should I move the objects ?". 

\subsubsection{Evaluation process}
In order to test the generalization capabilities of the different representations, we define 3 levels of generalization: 
\begin{itemize}
    \item (L1) \textit{unseen distractor colors}
    \item (L2) \textit{unseen background colors}
    \item (L3) \textit{novel distractor sizes}
\end{itemize}

The unseen distractors colors are randomly sampled from the following set : [\textit{\mycirc[clpink], \mycirc[clorange], \mycirc[cllime], \mycirc[clblack], \mycirc[cldgreen], \mycirc[cldblue], \mycirc[cldred]}] while the unseen set of the background is [\textit{\mycirc[clyellowbg], \mycirc[clpinkbg], \mycirc[clcyanbg], \mycirc[clorange], \mycirc[cldgreenbg]}]. The table color is randomized between black and white.

The metric we use for evaluation is the mean success rate over 100 different rollouts, with a new seed for each rollout. We repeat the experiment 3 times (300 trajectories) to compute then mean and standard deviation, which are presented in the following tables.

\subsubsection{Baselines}
As we aim to evaluate whether the structure provided by an object-centric representation is beneficial for learning downstream robotic tasks, we decided to use two global reprsentation models as baselines: one classical CV model and one robotic pre-trained representation. Specifically, we use the pre-trained vision transformer (ViT) version of DINO \cite{caron2021emergingpropertiesselfsupervisedvision}  and the R3M \cite{nair2022r3muniversalvisualrepresentation} model. For the object-centric model, we tested the original version of SAVi \cite{kipf2022conditionalobjectcentriclearningvideo}. All models are frozen during policy learning.

\subsection{Quantitative Results}
In the following, we refer to the policy models by the encoder used in their behavior learning process : R3M, DINO or SAVi.

In the task \textit{Push Cube to target} presented in Table \ref{push_cube}, the pre-trained representation of R3M performs the best. The task is relatively simple, as the primary requirement is to understand the relative position of the cube to the target in order to push or pull it to the correct target spot. The DINO model fails to perform any successful rollouts. We hypothesize that, even though the task is simple, the representation model may easily overfit to the background or the table, as it dominates the input observation. R3M overcomes this issue by learning dynamic information through its time-contrastive learning, enabling it to use this knowledge to complete the task. The object-centric model SAVi is able to perform the task but with a lower success-rate than R3M. 

To better understand these results, we visualized the different rollouts produced by the models : R3M demonstrates the ability to recover from suboptimal trajectories where the robot initially fails to push the cube to the target, a capability not observed in the object-centric model, which tends to get stuck in such situations. We hypothesize that while the object-centric representation is beneficial, it may also require dynamic information during the learning process. However, it is important to note that our model was directly trained on the final data without requiring any large-scale pre-training on out-of-domain data and operates with a significantly smaller scale (approximately 25 times fewer parameters). This suggests that larger pre-training and a more complex loss function could potentially yield even better results.

We also observe an interesting trend in the \textit{L1} scenario: R3M experiences a significant drop in success rate (approximately 60\%). This scenario introduces distractor colors unseen during training. In contrast, the object-centric method is notably robust in this scenario, with only a 14\% drop in success rate. Since the encoder segments the objects in the scene into separate slots, it can isolate the object of interest to complete the task while avoiding reliance on "noisy" information. A similar trend is observed in other generalization scenarios: the object-centric method consistently demonstrates robustness to out-of-distribution scenarios, experiencing relatively smaller performance drops compared to R3M.

\begin{table}[h]
\caption{Success rate of \textit{Push Cube to target} task over the different generalization scenarios}
\label{push_cube}
\begin{center}
\begin{tabular}{|c||c|c|c|}
\hline
 & DINO & R3M & SAVi \\
 \hline
$\emptyset$ & 0.  & \textbf{0.88} $\pm$ 0.07  & 0.69 $\pm$ 0.06 \\
\hline
L1 & 0.  & 0.26 $\pm$ 0.06  & \textbf{0.55} $\pm$ 0.02  \\
\hline
L2 & 0.  & \textbf{0.70} $\pm$ 0.01 & 0.60 $\pm$ 0.04   \\
\hline
L3 & 0.  & \textbf{0.76} $\pm$ 0.01 & 0.56 $\pm$ 0.08 \\
\hline
\end{tabular}
\end{center}
\end{table}

The task \textit{Pick cube}, presented in Table \ref{pick_cube}, is more complex as it requires the model to grasp the red cube and position it at a specific relative location above. This task demands greater precision and reasoning to identify the correct shape in the scene and place it in the correct position.

The results in the Table \ref{pick_cube} show that the object-centric model outperforms the other models. As expected, the DINO model is unable to perform this task, similar to the previous one. Surprisingly, R3M also fails in this task, but not as we expected. While R3M is able to grasp the cube in most rollouts, the single front-view input and lack of additional information make it difficult for the model to correctly position the cube. As a result, it often releases the cube in the wrong place, leading to failure in most cases. Although the dynamic information encoded in R3M helps the model approach and grasp the cube, it is insufficient for accurately placing the cube in space, as no visual cue for the target position is provided.

The object-centric model successfully performs the task, with only a small drop in performance compared to the first task. Notably, it is the only model capable of completing this task. Furthermore, the object-centric model demonstrates robustness across different generalization scenarios, with an approximate performance drop of just 10\% per scenario.

\begin{table}[h]
\caption{Success rate of \textit{Pick cube} task over the different generalization scenarios}
\label{pick_cube}
\begin{center}
\begin{tabular}{|c||c|c|c|}
\hline
 & DINO & R3M  & SAVi \\
 \hline
$\emptyset$ & 0. & 0.  & \textbf{0.56} $\pm$ 0.01 \\
\hline
L1 & 0.  & 0.  & \textbf{0.44} $\pm$ 0.04 \\
\hline
L2 & 0.  & 0.  &  \textbf{0.48} $\pm$ 0.07 \\
\hline
L3 & 0.  & 0.  & \textbf{0.51} $\pm$ 0.02 \\
\hline
\end{tabular}
\end{center}
\end{table}

For the task \textit{Place cube in bin}, none of the models are able to achieve a positive success rate. The task requires the model to first grasp the cube, place it on top of a box and realease it. The model must understand which cube to grab, identify the target position, and determine how to move from one to the other.

\begin{figure}[h]
\begin{center}
\includegraphics[width=0.55\linewidth]{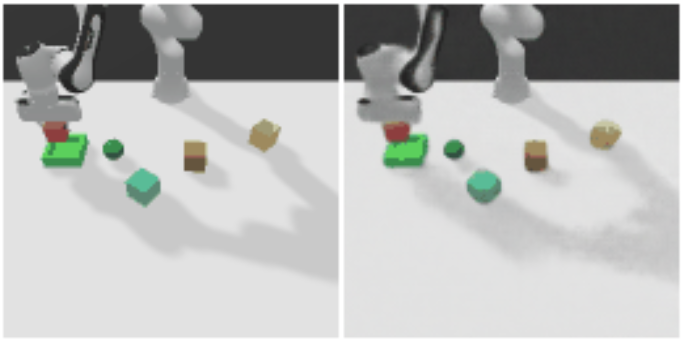}
\end{center}
\caption{\textbf{Failure case in the last task} : the robot is blocked on top of the green box and do not realease the cube. Left : Input observation; Right : Model reconstruction}
\label{fig:fail}
\end{figure}

Interestingly, both the R3M and SAVi models fail, yet they are able to grasp the cube and move it toward the box. However, the R3M model is unable to drop the cube inside the box, and instead, it falls next to it. This suggests a need for better spatial understanding. The SAVi-based model, on the other hand, is able to position itself directly above the box but do not drop the cube. We hypothesize that this is due to the uncertainty introduced by the single-view input. In many cases, the model appears to perceive the cube as already inside the box, as illustrated in Figure \ref{fig:fail}. We believe that using a mounted camera view would resolve this issue.

\begin{figure}[h]
\begin{center}
\includegraphics[width=0.55\linewidth]{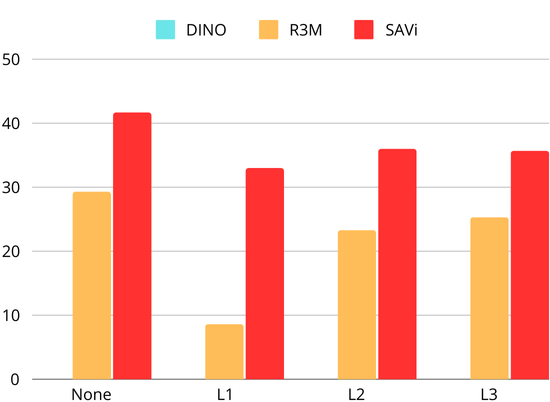}
\end{center}
\caption{\textbf{Overall success rate}: mean success rate over the 3 tasks on the different generalization scenarios}
\label{fig:compar}
\end{figure}

Finally, the Figure \ref{fig:compar} provides an overall comparison of the mean success rate across all scenarios in the different tasks for each model. Since the DINO and R3M models are unable to perform the second and third tasks of our benchmark, they overall underperform compared to the object-centric alternative, which demonstrates robustness across different generalization levels with only small drops in performance. 

The underperformance of R3M in the \textit{L1} scenario suggests the need to further explore further this direction: how robust are global representation models to unseen distractors ?

\subsection{Qualitative Results}
\begin{figure}[h]
\begin{center}
\includegraphics[width=\linewidth]{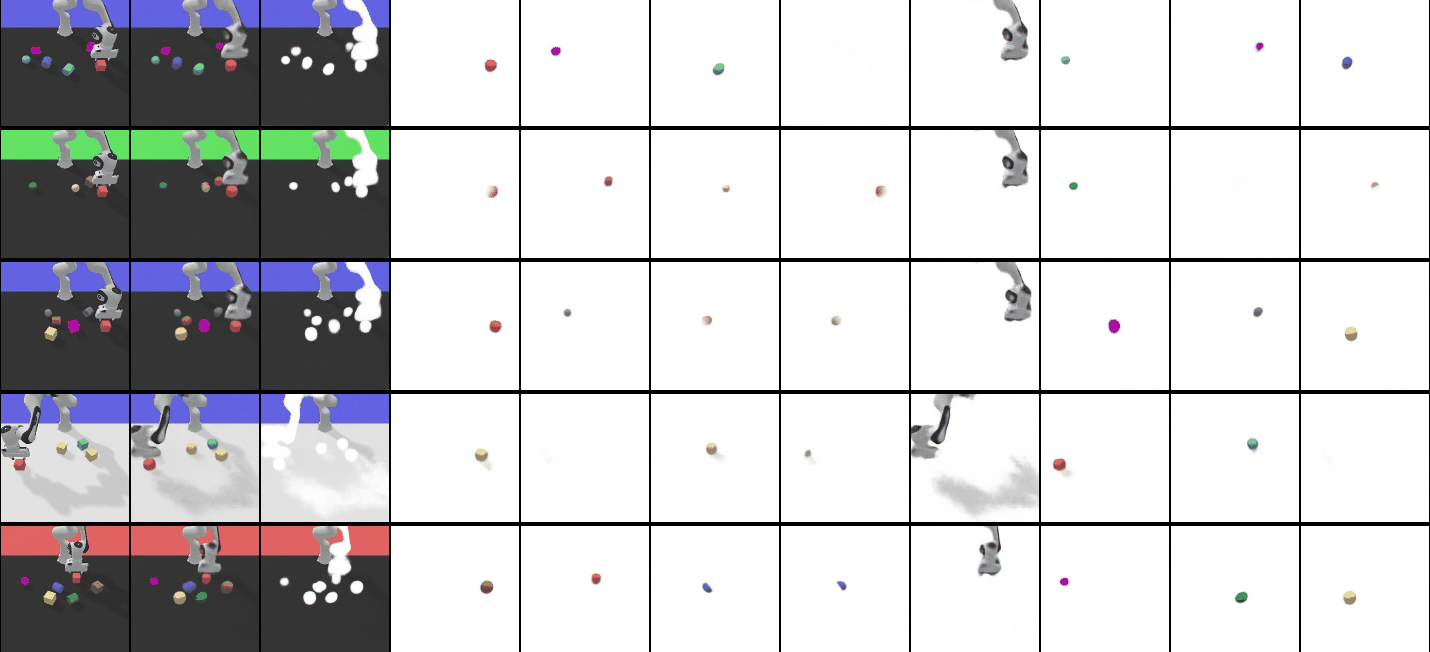}
\caption{\textbf{Object decomposition on rollout} From left to right : input image, full reconstruction, slot decomposition}
\label{fig:decomp}
\end{center}
\end{figure}

For validation purposes, we visualize the reconstruction of the predicted next slot by the object-centric model during behavior learning, comparing it to the rollout obtained in the real environment with the predicted actions. Figure \ref{fig:decomp} shows the decomposition for five different rollouts. It demonstrates that the model accurately splits the input observation into distinct parts: the background, the robot, and the various objects in the scene. By leveraging this pre-trained model, future exploration could focus on the concept of world models \cite{hawm, hafner2020dreamcontrollearningbehaviors}, enabling "learning within an imagined world."

\section{Conclusion}
We introduced a new simple framework which combines behavior cloning with object-centric representations. To validate the effectiveness of our approach compared to existing global representation methods, we propose three new challenging robotic tasks that feature increasing levels of difficulty and varying degrees of generalization for each task.
Despite a limited model size and data scale, our method achieves success in complex tasks where state-of-the-art computer vision and vision-based robotic models fail.

Furthermore, we demonstrate that object-centric methods can mitigate the effects of out-of-distribution scenarios, such as handling new distractors. This highlights the potential of object-centric representation methods as a promising direction for advancing the generalization of robotic manipulation learning. We hope this work inspires others to explore the use of object-centric representations in robotics.

\subsubsection{Limitations and future works}
Our work is currently limited to simulation settings involving relatively simple tasks. We plan to extend this research by exploring more complex environments with objects from everyday life. 

Additionally, we aim to scale our method to real-world applications by pretraining our object-centric approach on existing large-scale robotic datasets \cite{embodimentcollaboration2024openxembodimentroboticlearning}, leveraging recent state-of-the-art object-centric methods \cite{zadaianchuk2023objectcentriclearningrealworldvideos}. Finally, we seek to have a deeper understanding of how different representations influence learning by conducting comparisons with a broader range of methods, including both object-centric and global approaches.

\subsubsection*{Acknowledgments}
This work was in part supported by the French national program of investment of the futur and the regions through the PSPC FAIR Waste project, as well as  the French Research Agency, l’Agence Nationale de Recherche (ANR), through the projects Chiron (ANR-20-IADJ-0001-01),  Aristotle (ANR-21-FAI1-0009-01), and Astérix (ANR-23-EDIA-0002-001).

\bibliographystyle{splncs04}
\bibliography{biblio}

\end{document}